\documentclass{isprs} % isprs class modified 23-04-2019 (Dennis Wittich)
\usepackage{subfigure}
\usepackage{setspace}
\usepackage{geometry} % added 27-02-2014 Markus Englich
\usepackage{epstopdf}
\usepackage[labelsep=period]{caption}  % added 14-04-2016 Markus Englich - Recommendation by Sebastian Brocks
\usepackage[british]{babel} 
\usepackage[hang]{footmisc}
\usepackage{enumitem}
\usepackage{fontawesome5}
\usepackage{booktabs}  
\usepackage{wasysym}
\usepackage{amssymb}

\usepackage{hyperref}

\usepackage[table]{xcolor}
\usepackage{multirow}
\usepackage{tabularx}
\usepackage{xstring}
\usepackage{pgfplots}
\usepackage{tikz}
\usepackage{tikz-3dplot}
\usepackage{pgf-pie}
\usetikzlibrary{calc, shapes, patterns, shadings, positioning, arrows.meta,3d, intersections}
\usepgfplotslibrary{fillbetween,colormaps}
\pgfplotsset{compat=1.17}

\newcommand{\High}{\textcolor{red!95}{\faCircle}}

\newcommand{\All}{\textcolor{green}{\faCircle}}
\newcommand{\Nul}{\Circle}

\newcommand{\HighI}{\textcolor{brown!75!red!70!blue}{\faCircle}}
\newcommand{\highI}{\textcolor{red!95!yellow}{\faCircle}}
\newcommand{\medI}{\textcolor{red!30!yellow}{\faCircle}}
\newcommand{\lowI}{\textcolor{yellow}{\faCircle}}
\newcommand{\LowI}{\textcolor{yellow!30!white}{\faCircle}}
\newcommand{\HighU}{\begin{tikzpicture}\draw[ultra thick,green!50!white,fill=white] (0,0) circle (1.3ex);\draw[white,fill=brown!75!red!70!blue] (0,0) circle (0.9ex);\end{tikzpicture}}
\newcommand{\HighV}{\begin{tikzpicture}\draw[ultra thick,red!70!black,fill=white] (0,0) circle (1.3ex);\draw[white,fill=brown!75!red!70!blue] (0,0) circle (0.9ex);\end{tikzpicture}}
\newcommand{\HighMo}{\begin{tikzpicture}\draw[ultra thick,cyan!40!white,fill=white] (0,0) circle (1.3ex);\draw[white,fill=brown!75!red!70!blue] (0,0) circle (0.9ex);\end{tikzpicture}}
\newcommand{\HighMa}{\begin{tikzpicture}\draw[ultra thick,blue!90!white,fill=white] (0,0) circle (1.3ex);\draw[white,fill=brown!75!red!70!blue] (0,0) circle (0.9ex);\end{tikzpicture}}
\newcommand{\highU}{\begin{tikzpicture}\draw[ultra thick,green!50!white,fill=white] (0,0) circle (1.3ex);\draw[white,fill=red!95!yellow] (0,0) circle (0.9ex);\end{tikzpicture}}
\newcommand{\highV}{\begin{tikzpicture}\draw[ultra thick,red!70!black,fill=white] (0,0) circle (1.3ex);\draw[white,fill=red!95!yellow] (0,0) circle (0.9ex);\end{tikzpicture}}
\newcommand{\highMa}{\begin{tikzpicture}\draw[ultra thick,blue!90!white,fill=white] (0,0) circle (1.3ex);\draw[white,fill=red!95!yellow] (0,0) circle (0.9ex);\end{tikzpicture}}
\newcommand{\highMo}{\begin{tikzpicture}\draw[ultra thick,cyan!40!white,fill=white] (0,0) circle (1.3ex);\draw[white,fill=red!95!yellow] (0,0) circle (0.9ex);\end{tikzpicture}}
\newcommand{\medU}{\begin{tikzpicture}\draw[ultra thick,green!50!white,fill=white] (0,0) circle (1.3ex);\draw[white,fill=red!30!yellow] (0,0) circle (0.9ex);\end{tikzpicture}}
\newcommand{\medV}{\begin{tikzpicture}\draw[ultra thick,red!70!black,fill=white] (0,0) circle (1.3ex);\draw[white,fill=red!30!yellow] (0,0) circle (0.9ex);\end{tikzpicture}}
\newcommand{\medMo}{\begin{tikzpicture}\draw[ultra thick,cyan!40!white,fill=white] (0,0) circle (1.3ex);\draw[white,fill=red!30!yellow] (0,0) circle (0.9ex);\end{tikzpicture}}
\newcommand{\medMa}{\begin{tikzpicture}\draw[ultra thick,blue!90!white,fill=white] (0,0) circle (1.3ex);\draw[white,fill=red!30!yellow] (0,0) circle (0.9ex);\end{tikzpicture}}
\newcommand{\lowU}{\begin{tikzpicture}\draw[ultra thick,green!50!white,fill=white] (0,0) circle (1.3ex);\draw[white,fill=yellow] (0,0) circle (0.9ex);\end{tikzpicture}}
\newcommand{\lowV}{\begin{tikzpicture}\draw[ultra thick,red!70!black,fill=white] (0,0) circle (1.3ex);\draw[white,fill=yellow] (0,0) circle (0.9ex);\end{tikzpicture}}
\newcommand{\lowMo}{\begin{tikzpicture}\draw[ultra thick,cyan!40!white,fill=white] (0,0) circle (1.3ex);\draw[white,fill=yellow] (0,0) circle (0.9ex);\end{tikzpicture}}
\newcommand{\lowMa}{\begin{tikzpicture}\draw[ultra thick,blue!90!white,fill=white] (0,0) circle (1.3ex);\draw[white,fill=yellow] (0,0) circle (0.9ex);\end{tikzpicture}}
\newcommand{\LowU}{\begin{tikzpicture}\draw[ultra thick,green!50!white,fill=white] (0,0) circle (1.3ex);\draw[white,fill=yellow!30!white] (0,0) circle (0.9ex);\end{tikzpicture}}
\newcommand{\LowV}{\begin{tikzpicture}\draw[ultra thick,red!70!black,fill=white] (0,0) circle (1.3ex);\draw[white,fill=yellow!30!white] (0,0) circle (0.9ex);\end{tikzpicture}}
\newcommand{\LowMo}{\begin{tikzpicture}\draw[ultra thick,cyan!40!white,fill=white] (0,0) circle (1.3ex);\draw[white,fill=yellow!30!white] (0,0) circle (0.9ex);\end{tikzpicture}}
\newcommand{\LowMa}{\begin{tikzpicture}\draw[ultra thick,blue!90!white,fill=white] (0,0) circle (1.3ex);\draw[white,fill=yellow!30!white] (0,0) circle (0.9ex);\end{tikzpicture}}
\newcommand{\model}{\tikz\draw[ultra thick,cyan!40!white,fill=white] (0,0) circle (1.0ex);}
\newcommand{\update}{\tikz\draw[ultra thick,green!50!white,fill=white] (0,0) circle (1.0ex);}
\newcommand{\validate}{\tikz\draw[ultra thick,red!70!black,fill=white] (0,0) circle (1.0ex);}
\newcommand{\mapping}{\tikz\draw[ultra thick,blue!90!white,fill=white] (0,0) circle (1.0ex);}
\newcommand{\EG}{\textit{e.g.,~}}
\newcommand{\IE}{\textit{i.e.,~}}

\begin{document}

\title{The Global-Local loop: what is missing in bridging the gap\\ between geospatial data from numerous communities?}
\date{}

% KAO: Remove extra spacing

% Anonymous submissions, authors' names should not be visible
% \author{
%  Orhan Altan\textsuperscript{1}, Ian Dowman\textsuperscript{2}, Florent Lafarge\textsuperscript{3}, Clément Mallet\textsuperscript{4}, Christian Heipke\textsuperscript{5} }
\author{Clément Mallet, Ana-Maria Raimond}

% KAO: Remove extra newline
% Anonymous submissions, authors' affiliations should not be visible
%\address{
%	\textsuperscript{1 }ITU, Civil Engineering Faculty, 80626 Maslak Istanbul, Turkey - (oaltan, tozg, kulur, seker)@itu.edu.tr\\
%	\textsuperscript{2 }Dept.\ of Geomatic Engineering, University College London, Gower Street, London, WC1E 6BT UK - idowman@ge.ucl.ac.uk\\
%	\textsuperscript{3 }Université Côte d’Azur, INRIA – Sophia-Antipolis, France – florent.lafarge@inria.fr\\
%	\textsuperscript{4 }Univ. Gustave Eiffel, IGN-ENSG, LaSTIG – Saint-Mandé, France – clement.mallet@ign.fr\\
%	\textsuperscript{5 }Institute of Photogrammetry and GeoInformation, Leibniz Universit\"at Hannover, Germany - heipke@ipi.uni-hannover.de\\
%}
\address{Univ Gustave Eiffel, Géodata Paris, IGN, LASTIG, F-77454 Marne-la-Vallée, France\\
{firstname.lastname}@ign.fr}

% If the corresponding author is NOT the final author, always add a % space before the subsequent comma, i.e.
% first author name\textsuperscript{a,}\thanks{Corresponding author} , % second author name \textsuperscript{b}, etc.
% thanks to Niclas Borlin 05-05-2016
% information on the corresponding author should not be used any longer and has been commented out
% C. Heipke, Jan 03,2024

% the use of the information of commissions and working groups should not be used any longer and has been commented out
% C. Heipke, Sept. 20,2022
%\commission{XX, }{YY} %This field is optional. If filled, XX and YY should be replaced by adequate numbers. See https://www2.isprs.org/commissions/
%\workinggroup{XX/YY} %This field is optional.
%\icwg{}   %This field is optional.

% KAO: Use times symbol
\abstract{
We face a unprecedented amount of geospatial data, describing directly or indirectly the Earth Surface at multiple spatial, temporal, and semantic scales, and stemming from numerous contributors, from satellites to citizens. The main challenge in all the geospatial-related communities lies in suitably leveraging a combination of some of the sources for either a generic or a thematic application. Certain data fusion schemes are predominantly exploited: they correspond to popular tasks with mainstream data sources, \EG free archives of Sentinel images coupled with OpenStreetMap data under an open and widespread deep-learning backbone for land-cover mapping purposes. Most of these approaches unfortunately operate under a "master-slave" paradigm, where one source is basically integrated to help processing the "main" source, without mutual advantages (\EG large-scale estimation of a given biophysical variable using in-situ observations) and under a specific community bias.
We argue that numerous key data fusion configurations, and in particular the effort in symmetrizing the exploitation of multiple data sources, are insufficiently addressed while being highly beneficial for generic or thematic applications. Bridges and retroactions between scales, communities and their respective sources are lacking, neglecting the utmost potential of such a "\textit{global-local loop}".  In this paper, we propose to establish the most relevant interaction schemes through illustrative use cases. We subsequently discuss under-explored research directions that could take advantage of leveraging available data through multiples extents and communities. 

}


\maketitle

%%%%%%%%%%%%%%%%%%%%%%%%%%%%%%%%%%%%%%%%%%%%%%%%%%%%%%%%%%%

\section{Introduction}\label{sec:introduction}
 
% KAO: Sloppy spacing ensures non-overfull lines. Can be removed if this is not an issue.
\sloppy
In the last decade, the photogrammetric, remote sensing and geographical information sciences communities have witnessed a sharp increase in initiatives leveraging data of multiple sources, formats, and resolutions to address either generic or specific geospatial applications (\EG land-cover mapping at national scale or tree disease monitoring in a city, respectively). Such opportunities stem for the deluge of complementary data sources at multiple extents, that can also be either generic or thematically driven, leading to an extremely multi-modal world \cite{DevisAI}. Now, we have access to multi-format sources (images, text, vector geodatabases, etc.) at multiple resolutions and scales, spanning over a large diversity of geographical extents: \EG social media correspond to irregularly sampled \textit{local} and \textit{thematic} information covering a \textit{global} extent, while a national land-cover database corresponds to a more \textit{regional} and \textit{generic} information with dense coverage. Such diversity offers unprecedented opportunities while requiring solving significant methodological challenges. This explains why, in our open-data era, the high relevance and heterogeneity of such data for numerous applications has also stimulated research in side communities such as computer vision and machine learning \cite{Dimitrovski} \cite{NEURIPS2023_4786c0d1} \cite{brown2025alphaearth}.\\
This has steadily increased the dialogue between disciplines to foster complementary interactions between their "respective" data sources (sources they produce, data they acquire or that they are used to work with), and their respective expectations \cite{STAMMLERGOSSMANN2024101103}. It is now assumed that addressing key social and environmental applications and  methodological findings require a strong and educated interplay between disciplines \cite{PlosOne_longterm} \cite{Surprising}. %the three levels in order to handle the four main processes of the lifecycle of geospatial data: modelling, mapping, validation, and update. A vast body of literature exists to illustrate each of these four stages (defined in Section~\ref{subsec:types_interactions}).

Despite such a momentum, two major limitations can be noted in most initiatives involving geospatial data.
%\begin{itemize}[leftmargin=*]
    %\item
\paragraph{(1) Insufficient and asymmetric exploitation of the various categories of data sources.} Such exploitation depends on the application and the domain the contributors fall into. This is a community-based bias, where every community defines what is "auxiliary source" or "metadata" \cite{Wang_2025_ICCV}. Each geographical extent and data source has its specific contribution but many approaches limit themselves to two of them and ignore the other ones. This narrows down their potential. It often comes from the fact that one spatial extent is often a \textit{master} extent, from which the core information is extracted. The \textit{slave}  one(s) is(are) only inserted for a specific step (supervision or alternative input source), without (i) being appropriately handled and (ii) retroaction loop, for the benefit of this \textit{slave} extent. For instance, we train or fine-tune an EO-based global deep learning land-cover model for optimizing the performance on a validation or a test set (with samples irregularly sampling the Earth surface) or for minimizing the supervision regime. However, the inherent bias of the sources is not questioned \cite{BIRD2014144}, and upscaling the model to unseen areas may neglect (i) the validation of other \textit{in situ} data sources available for the classes of interest and (ii) the heterogeneity of the input images when performing quality assessment of the results (here, \textit{master}: global, \textit{slave}: local  \cite{PAPOUTSIS2023250}). This is often cast as a "model-centric" perspective instead of the "data-centric" one \cite{baek2025aisensebetterjust} \cite{Ribana_GRSM25}.
%    \item
\paragraph{(2) Limited two-way interactions.} Few initiatives adopt a back-and-forth or \textit{retroaction} strategy. Slave level(s)  can help improving the \textit{master} one in terms of spatial, temporal, semantic information: \EG training the land-cover semantic segmentation model with geodatabases without handling their necessary update (missing link: \textit{global} $\rightarrow$ \textit{local} \cite{FRITZ2012110} \cite{LI202166}) or with text without questioning how informative or biased it is \cite{Daroya_2025_ICCV} \cite{TEOChat}. This can be cast into the "lifecycle data assessment" or "critical analysis of the sources" paradigms \cite{Nature_LCA}.

%   \end{itemize}
Such gaps jointly stem from methodological challenges and distinct practices between communities (\EG method-driven, data-driven, social human vs environmental vs computer sciences) and disciplines \cite{Nature_historical_text}. Such a siloed reasoning prevents holistic approaches \cite{Edito_Silos} and fully interoperable pipelines \cite{EarthAI}. In this paper, we discuss such limitations through a description of the main categories and geographical extents of geospatial data (Section~\ref{sec:sources}), before showcasing how they interact each other sources through main use cases, and defining subsequent challenges (Section~\ref{sec:challenges}). This eventually leads to present major lines of research that have been obviated or barely investigated so far (Section~\ref{sec:perspectives}).

%In parallel, there is also a huge demand for local scale quantification and qualification of land surfaces, which is today mainly fulfilled by in-situ measurements and observatories, through institutional and/or citizen science campaigns. Such campaigns, now even coupled with prompts when integrating LLMs or VLMs, are extensively carried out and offer a complementary knowledge to coarser global land-cover/land-use maps.

%\section{Problem statement}

%
%%%%%%%%%%%%%%%%%%%%%%%%%%%%%%%%%%%%%%%%%%%%%%%%%%%%%%%%%%%
%%%%%%%%%%%%%%%%%%%%%%%%%%%%%%%%%%%%%%%%%%%%%%%%%%%%%%%%%%%
\section{Data sources}\label{sec:sources}

\subsection{How to categorize geospatial data?}
\begin{figure*}[ht!]
\begin{center}
		\includegraphics[width=1.0\columnwidth]{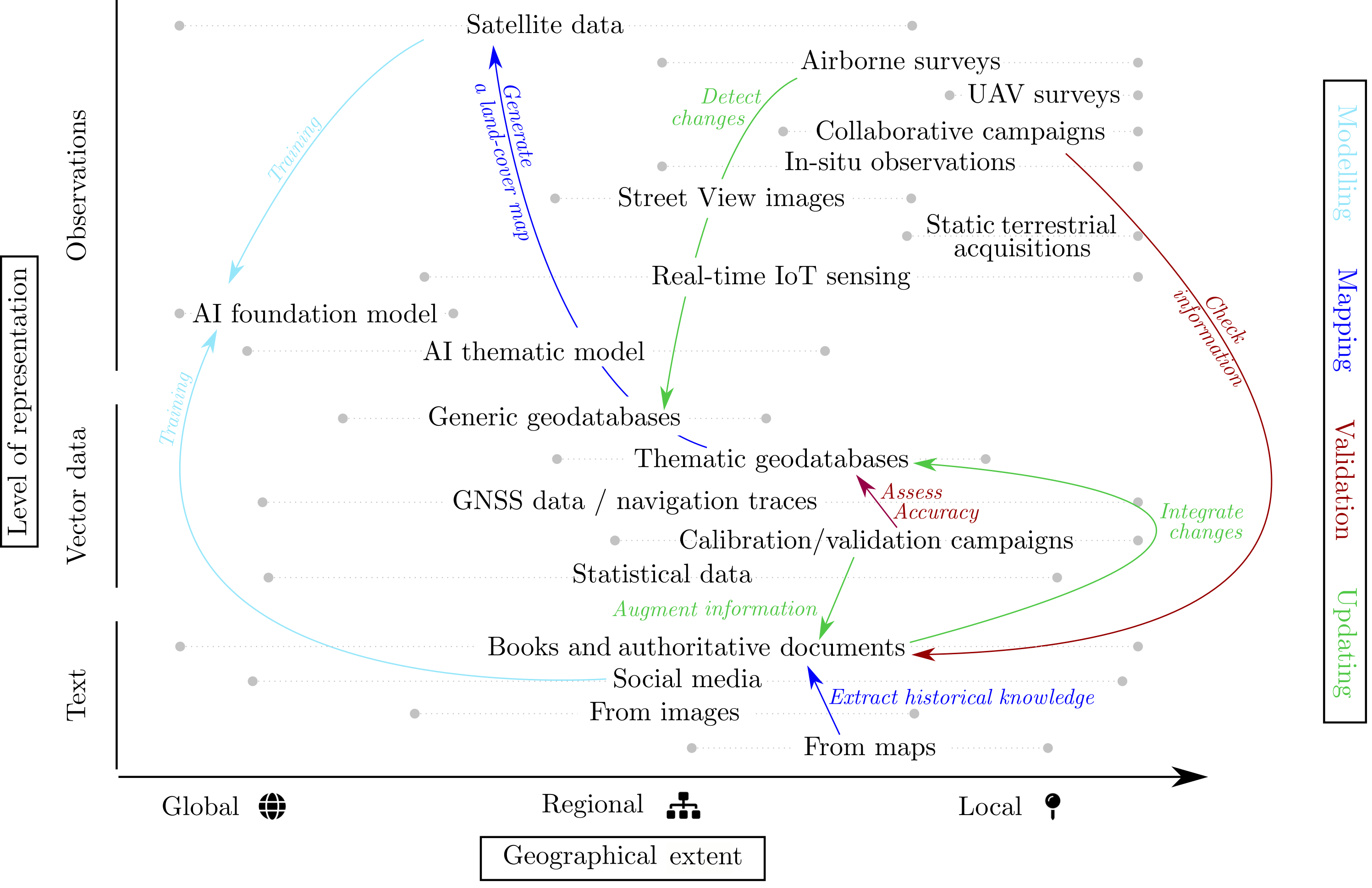}
	\caption{Main geospatial data sources and their most widespread one-way interactions (one per pair of levels of representations). All conceivable possibilities are not displayed for the sake of simplicity.\\
    \ldots\texttt{source}\ldots\;\; indicates how a \texttt{source} spans over the three geographical extents. 
    \texttt{slave}\:$\curvearrowright$\:\texttt{master} showcases a \textit{use case} between a \texttt{slave} and a \texttt{master} source, coloured by the lifecycle stage it corresponds to.}
\label{fig:figure_usecases}
\end{center}
\end{figure*}

Here, we define two main dimensions to categorize and present the main geospatial data sources.
\begin{itemize}[leftmargin=*]
    \item \textbf{The geographical extent}: the physical area over which the source spans (Section~\ref{subsec:geo_extent}). It is preferred to the concept of \textit{scale} or \textit{spatial level}. We believe it better captures the discrepancies and complementarities between sources, while being less confusing between communities.
    \item \textbf{The level of representation}: to jointly assess how close for the physical world we are in terms of measurement, structuration or geolocation  (Section~\ref{subsec:levels_rep}). It is often simplified with the concepts of \textit{format} or even \textit{source}.
\end{itemize}
Geospatial data is often clustered and presented according to the various resolutions: spatial, temporal, spectral, and semantics. However, this dimension does not sufficiently point out the diversity between communities and offers a fuzzy snapshot of the situation. The term "modality" is also often adopted in some communities but discarded here. We have noticed significant discrepancies on what is a "modality" in the literature, especially between communities (\EG Geographical Information Sciences, GIS, \cite{Chapuis02092022} vs computer vision \cite{Jakubik_2025_ICCV}), and how the clustering is performed.\\
In the following, we present both dimensions, without detailing all conceivable geospatial sources for each of the categories we define (the reader can refer to multiple review papers in the literature, \EG \cite{rs17030550}).\\
Table~\ref{tab:sources} provides a tentative quantitative summary of the significance of each source according to (i) the resolutions and (ii) how generic they are (\IE how far from a specific thematic application). Main sources are listed in Figure~\ref{fig:figure_usecases}.%, and their geographical extent specified.

\subsection{Geographical extent}
\label{subsec:geo_extent}
We define three main levels, assuming a given source can span over multiple levels, \EG when capturing a very local phenomenon but over large areas. One can note that the finer extent, the more educated and specific the information.
%\begin{itemize}[leftmargin=*]
%    \item
\paragraph{The Global extent} corresponds to sources densely available worldwide (\EG Earth Observation (EO) with numerous satellites providing daily global coverage at multiple spatial, temporal and spectral scales) or that sample the Earth surface over numerous locations (\EG navigation traces, Internet-of-Things -IoT- sensors). They can be either generic or thematic (books, authoritative documents and social media are good examples that gather both possibilities).\\ 
In parallel, EO data has stimulated researchers of various fields to develop machine learning models for numerous thematics and now the so-called foundation/world models \cite{FM_GRSM26} and agentic systems \cite{Mapeval_2025_ICML}, able to agnostically ingest any kind of EO data, social media information or navigation traces. This offers, at another processing level, a new generic input data for multiple downstream tasks. 
%    \item
\paragraph{At regional extents,}national surveys provide remotely sensed data with higher spatial resolution, still with dense coverage \cite{KISSLING2023108798}. They appropriately complement vector geodatabases, with higher semantics, statistical data (\EG population or weather), as well as observatories, that can help monitoring specific patterns \cite{population_plos_one}, objects and landscapes with high temporal revisit over a carefully designed sampling of large extents \cite{ign2025flairhub}.
\paragraph{At local extents,}in-situ sensors, images, traces or text from social media, field or aerial surveys as well as collaborative contributions through citizen science \cite{Zheng2018,Liu03042021} are pivotal to derive on-purpose information, adapted to specific users' requirements or environments \cite{Kitchin}.

%\end{itemize}
We consider continental-wide data at the \textit{global} extent, while national coverages at the \textit{regional} one. We assume that the \textit{regional} extent exhibits a certain spatial homogeneity (in terms of data coverage or producers), and therefore a lower complexity that the \textit{global} one. This explains this categorization.
\begin{table}[h]
%\footnotesize
	\centering
		\begin{tabular}{l c c c c c c c c c}
        \toprule
        Level of representation & \multicolumn{3}{c}{Observations}& \multicolumn{3}{c}{Vector data}& \multicolumn{3}{c}{Text}\\
        \cmidrule(lr){2-4}\cmidrule(lr){5-7}\cmidrule(lr){8-10}
        Extent &\faGlobe &\faSitemap &\faMapPin &\faGlobe &\faSitemap &\faMapPin &\faGlobe &\faSitemap &\faMapPin \\
        \midrule
        Spatial resolution &\medI &\highI &\HighI &\medI &\HighI &\HighI &\All &\medI &\HighI\\
        Temporal resolution &\HighI &\medI &\All &\highI &\lowI &\All &\highI &\lowI &\highI\\
        Semantics &\LowI &\LowI &\LowI &\lowI &\HighI &\All &\High &\HighI &\lowI\\
        Attributes &\medI &\highI &\HighI &\lowI &\medI &\HighI &\LowI &\LowI &\lowI\\
        \midrule
        Genericity & \HighI &\HighI &\lowI &\HighI &\medI &\LowI &\medI &\medI &\lowI \\
        \bottomrule
        &&&&&&&&&\\
		\end{tabular}
        \\
    \begin{tabular}{c r@{ : }l r@{ : }l}
\textbf{Extent} & \faGlobe & global & \faSitemap & regional\;\;\;\;\; \faMapPin\:: local \\
\textbf{Significance} &\HighI\highI\medI\lowI\LowI & high$\rightarrow$low & \All & high \& low 
\end{tabular}    
	\caption{Main categories of geospatial sources, assessed w.r.t. their significance in providing information according to (i) their resolutions, (ii) the availability of (additional) bio-geophysical attributes, and (iii) the genericity (\IE whether they have been produced or acquired for a specific purpose).  Such assessment is performed according to our personal evaluation. The spectral resolution is ignored for simplicity.\\
    \underline{Example}: \textit{a global land-cover database (vector) is designed to be generic, has a medium spatial resolution, but probably a high temporal one (since updated with EO sensors), few semantic information (limited nomenclature) and few attributes (\EG height or density of objects)}.}
\label{tab:sources}
\end{table}

%%%%%%%%%
\subsection{Levels of representation}

We classify the level of representation by distinguishing three types of input data: observations, vector data, and text. Our categorisation is based on the format of \textit{initial} acquisition process, rather than on the format or information issued from any transformations that may subsequently be applied to it (\textit{e.g.}, extract locations as point vector data from text, or segment an image into feature‑based vectors.

\label{subsec:levels_rep}
\subsubsection{Observations.}
This includes quantitative data of the Earth surface (and its objects), acquired by either space-based, airborne, or ground-based devices. We also integrate reanalysis data \EG from environmental models. The full range of spatial, spectral and temporal resolutions is available. The \textit{global} extent includes for instance Earth Observation satellite data, the \textit{regional} one large-scale or national surveys (\EG orthophotos, 3D lidar), while the \textit{local} one in-situ observatories, IoT networks, local field campaigns such as crowdsourced or UAV  measurements. Most of the time, they come without any semantics (see Table~\ref{tab:sources}). However, one may note that the smaller the extent, the less generic and the more thematic the observation. At the same time, an effort is made to gather and normalize local field data to propose global datasets sampling multiple places \cite{cherif2025hyperspectraldataset}.
Deep learning models (in particular world models) that have been trained to encode such observations, either with or without semantics, are also included in this category. 

%%%%%%%%%
\subsubsection{Vector data.}
This category gathers structured data that can be fed into geodatabases or already integrated into such databases, with a high level of semantics and often (complementary) attributes. Most of the time pre-processed, this no longer represents raw level data (which makes the main difference with \textit{Observations} that is most naturally structured using a vector model). Again, this ranges from authoritative to citizen-based sources, and from agnostic to thematic purposes.  
At \textit{global}, and \textit{regional} extents, this includes land-cover/land-use maps or any geolocated administrative or statistical data (\EG census, population data \cite{UTAZI2025100744}); at \textit{regional} and  
\textit{local} ones, navigation traces \cite{ijgi11020078}, 3D models and any local representation of a territory \cite{10.1145/3618108}. We integrate in such a category field data/ground truth information used for calibrating or validating statistical models tailored for Earth surface semantic description. 

%%%%%%%%%
\subsubsection{Text data.}
Any geolocated textual information, or that can be geolocated, \EG with Named Entity Recognition, is relevant, either for authoritative or informal sources \cite{Syed04052025}. It captures information lacking in the two other categories, especially at \textit{regional} and \textit{local} extents. This also brings a strong historical depth \cite{icdar_bench}, more direct contextual or relationship information compared with raw sensed observations, but with limited generalisation ability. Alternatively, at \textit{global} extents, social media bring a massive yet unstructured, local, and biased source of information, especially for non officially documented areas or topics \cite{ijgi11010019}. In parallel,
Language-Image Pre-training deep-based models and more specifically Vision-Language Models (VLM) have emerged as a trustworthy solution for numerous geospatial applications \cite{Danish_2025_ICCV}, even without georeferenced textual data. Again, the more global, the more generic.

%%%%%%%%%%%%%%%%%%%%%%%%%%%%%%%%%%%%%%%%%%%%%%%%%%%%%%%%%%%
%%%%%%%%%%%%%%%%%%%%%%%%%%%%%%%%%%%%%%%%%%%%%%%%%%%%%%%%%%%

\section{Current situation in geospatial data interactions}\label{sec:challenges}
The way several data sources are combined in our communities illustrate the current bias and missing links. Such combinations can be determined by (i) the underlying \textit{process} at stake (Section~\ref{subsec:types_interactions})  and (ii) the magnitude of such interactions (Section~\ref{subsec:level_interactions}). While the first step can be performed through more or less popular \textit{use cases}, the second one is only carried out based on our perception of the field. An exhaustive quantitative analysis could only be performed through a suitably formalized ontology-based framework \cite{Ontology}.\\
Here, we limit ourselves to pair-wise interactions both for the sake of simplicity and since few examples exist of higher interactions, apart from multi-modal foundation models (\textit{EO images + labels from geodatabases + pseudo-labels for VLMs}) and change integration frameworks in national mapping agencies (\textit{combination of alerts from EO change detection pipelines +  from end-users or collaborative campaigns + from authoritative documents}).
\begin{table*}[h]
	\centering
    \begin{tabular}{c |c| c c c| c c c| c c c}
\toprule
       \multicolumn{1}{c}{Level of representation} &\textit{Master}& \multicolumn{3}{c}{Observations}& \multicolumn{3}{c}{Vector data}& \multicolumn{3}{c}{Text}\\
        \cmidrule(lr){2-2}\cmidrule(lr){3-5}\cmidrule(lr){6-8}\cmidrule(lr){9-11}
         \textit{Slave}&Extent&\faGlobe &\faSitemap &\faMapPin &\faGlobe &\faSitemap &\faMapPin &\faGlobe &\faSitemap &\faMapPin \\
   \midrule
   \multirow{3}{*}{Observations} &\centering \faGlobe&\Nul&\HighMo&\medMo&\highMa&\medU&\medU&\HighMo&\medU&\LowMa\\
 &\centering\faSitemap&\HighMo&\Nul&\highMa&\medMa&\HighU&\medU&\lowMa&\HighV&\highMa\\
 &\centering\faMapPin&\highMo&\HighMa&\Nul&\highU&\highV&\HighMa&\lowMa&\medMo&\highU\\
 \midrule
 \multirow{3}{*}{Vector}&\centering\faGlobe&\HighMa&\medV&\lowV&\Nul&\lowU&\LowMa&\HighMa&\highV&\lowV\\
 &\centering\faSitemap&\highMa&\HighV&\highV&\highMo&\Nul&\medMa&\lowMa&\highV&\lowV\\
 &\centering\faMapPin&\highV&\highU&\HighMa&\medV&\HighV&\Nul&\LowMa&\medU&\HighU\\
 \midrule
 \multirow{3}{*}{Text}&\centering\faGlobe&\highMo&\lowMo&\LowV&\medMo&\LowMo&\LowV&\Nul&\medU&\LowU\\
 &\centering\faSitemap&\lowV&\medMo&\lowU&\LowV&\highU&\lowU&\medV&\Nul&\medV\\
 &\centering\faMapPin&\LowV&\lowV&\highU&\LowV&\highU&\HighU&\medU&\highMa&\Nul\\
 \bottomrule
\end{tabular}\\
\begin{tabular}{c r@{ : }l r@{ : }l}
\textbf{Extent} & \faGlobe & global & \faSitemap & regional\;\;\;\;\; \faGlobe\:: local
\end{tabular} |
\begin{tabular}{c r@{ : }l r@{ : }l }
\textbf{Significance} &\HighI\highI\medI\lowI\LowI & high~$\rightarrow$~low  &  \Nul & non assessed
\end{tabular}   \\
\begin{tabular}{c r@{ : }l r@{ : }l r@{ : }l r@{ : }l}
\textbf{Lifecycle stage} & \model & modelling & \mapping & mapping & \validate & validation & \update & updating
\end{tabular}

	\caption{Current level of interactions between the three main categories of geospatial data, w.r.t their geographical extent (pair-wise analysis only). \textit{Master} indicates the primary exploited source, that is combined with a \textit{slave} one for carrying out one of the four processes identified as critical in the data lifecycle. We therefore also specify which process is prominently addressed in each interaction. \textbf{An interactive version, detailing use cases for each case is available at \href{https://www.umr-lastig.fr/global-local/}{https://www.umr-lastig.fr/global-local/}.}}
\label{tab:interactions}
\end{table*}

\subsection{Which interaction cases today?}
\label{subsec:types_interactions}
%Geospatial data analysis is henceforth natively multi-scale, multi-format, multi-resolutions, often termed as "\textit{multi-modal}". Various sources are jointly exploited with growing diversity. The aim \anamaria{l'objectif de qui? D'une communauté? d'un recherche} is to collect complementary information from the increasing number of (open) datasets made available by most communities and public authorities.
Main interactions are categorized according to the main stages that exist in the lifecycle of geospatial data. We illustrate each stage with the main use cases that exist in the communities.\\
In Table~\ref{tab:interactions}, we provide the most significant process for each conceivable pair of data sources. In Figure~\ref{fig:figure_usecases}, we specify only the most popular use case for each pair of levels of representation.

For this analysis, we retain the following four main stages.
\begin{itemize}[leftmargin=*]
    \item \textbf{Modelling}: It describes the process that turns a given source into a more informative level: semantization, attribute estimation, noise removal, enriched encoding (for Artificial Intelligence - AI - models). It can cover  the calibration of rule-based, statistical, and physical models but not their inference. The term "\textit{analysis}" is alternatively employed in the literature. The main use cases are:
    \begin{itemize}[leftmargin=*]
        \item[\faAngleRight] Indexing for data retrieval or matching, necessary for large archives, in particular those including non georeferenced documents (\EG images and texts from Galleries, Libraries, Archives and Museums).
        \item[\faAngleRight] Integration, for interoperability, fusion, cataloguing multiple sources. It includes normalization, standardization, or ontology-based solutions. It is all the more necessary when coupling data from different producers/contributors or distinct spatial supports (\EG statistical or forest inventory data may not represent fully localized information).
        \item[\faAngleRight] Training a machine learning (ML) model, in particular, in recent years AI (world or thematic) models, using ground truth data from almost any other source. The level of supervision depends on the amount of available reference data for the task at hand: discrete (\EG semantic segmentation) to continuous (\EG regression) labels.
        \item[\faAngleRight] Spatio-temporal planning or optimization, such as transportation optimization, accessibility, urban planning, etc.
    \end{itemize}
    \item \textbf{Mapping}: The process that applies a model or any other information extraction process to derive a geolocalized knowledge (not the case beforehand). The identified use cases are: %  that allows the production of geospatial data (e.g., producing LULC data, producing building footprints, producing Point Of Interest from text). 
     \begin{itemize}[leftmargin=*]
        \item[\faAngleRight] Information extraction from un-or mis-registered data sources: co-registration of two sources, or registration of one of the two sources, 3D information retrieval, up- and down-sampling (temporal, spatial, spectral), geolocating text from maps or images.
        \item[\faAngleRight] Inferring or fine-tuning a ML model for a particular task, for retrieving again either discrete (land-cover/land-use maps) or continuous variables (bio-geophysical parameter estimation) or both (object detection), as well as image descriptions (captioning).
    \end{itemize}
    \item \textbf{Validation}: The process that either measures whether the geospatial data fits the expected specifications (\EG according to a given stakeholder) or quantifies the performance of the modelling or the mapping stage. The main use cases therefore involve:
     \begin{itemize}[leftmargin=*]
        \item[\faAngleRight] Quality assessment, \EG of an observation (noise? correct registration?), of vector data (does the geodatabase meet the expected thematic or geometric accuracy?), of text data (does the information included in an authoritative document correspond to an object in the real-world?)
        \item[\faAngleRight] Performance assessment of data sources or of a machine learning model, here mainly based on reference data that can stem either from the methodological communities (computer vision, or machine learning), or a thematic one, for specific environmental tasks.
        \item[\faAngleRight] Accuracy assessment process design: protocols defining how to suitably make such assessment with reference data, in particular using external knowledge (\EG stratified sampling w.r.t. a thematic map). Such use case can be similarly adopted for training a machine learning model ("Modelling" process). The reader can refer to \cite{OLOFSSON201442} for an example in the change detection case.
    \end{itemize}
    \item \textbf{Updating}: It describes the process that integrates timely and reliable information into existing geospatial data, in particular geodatabases \cite{ZHANG201842}. This mainly corresponds to:
    %This includes, for example, change detection (e.g., detecting building changes from EO images or vector data at two dates t, t+1), change integration to produce a new snapshop at t+1 of building footprints by updating the existing snapshot at t with the detected changes, and remapping (i.e. producing from scratch a new snapshop  of building footprints at t+1 by using by using classification models. 
     \begin{itemize}[leftmargin=*]
        \item[\faAngleRight] Change detection from EO images, thematic vector data, or crowdsourced campaigns; 
        \item[\faAngleRight] Change integration in vector geodatabases: it can be inserting new objects or new attributes of existing objects ("\textit{Augment information}" in Figure~\ref{fig:figure_usecases}, \EG building heights from RADAR data or from town planning documents). 
    \end{itemize}
\end{itemize}
We voluntarily neglected several key stages of the lifecycle as well as challenges related to interconnections between sources and let the reader refer \EG to \cite{ijgi9020095}.\\

\subsection{Which level of interactions today?}
\label{subsec:level_interactions}
Table~\ref{tab:interactions} details our perception of the  current level of interactions between the three levels of interactions over the three proposed geographical extents. As presented in Section~\ref{sec:introduction}, we split between \textit{Master} and \textit{Slave} sources to reflect community or discipline bias toward its primary source of information. This results in a non-symmetrical description. Only pair-wise analysis is provided, our goal is mainly to roughly exacerbate the current shortcomings. On one hand, we subjectively evaluated how significantly each interaction is addressed in the literature. No meta-analysis was performed since we found particularly difficult and time-consuming to suitably harvest and analyse such an amount of papers in multiple communities with very different vocabularies (the use of ontology-based representations would allow a more formal description of processes and enable the derivation of quantitative indicators but was left to a subsequent study). On the other hand, we also tried to identify which process is primarily handled when coupling two specific sources. \\
One can first see that Table~\ref{tab:interactions} is rarely symmetric (for both aspects), which shows the one-way interactions. The tightest links correspond to "Observations~$\rightarrow$~Observations" and "Observations~$\rightarrow$~Vector", which mainly correspond to the massive effort in multi-modal EO image analysis, leveraging additional image sources or vector data to foster information and semantic extraction, respectively. The weakest links mainly navigate around text data ("Vector~$\rightarrow$~Text" and "Text~$\rightarrow$~Text"). It still remains under-exploited and considered as a "\textit{slave}" source in the computer sciences communities \cite{10049246} \cite{Mast14022025}, apart from social media data exploitation and interactive Vision-Language Models \cite{Soni_2025_CVPR}. The historical depth of existence of the sources of their adoption in a given community can also been noticed when comparing the nine main blocks of Table~\ref{tab:interactions}. "Observations and "Vector" are popular \textit{slave} sources since, respectively, many EO images have been made available and authoritative bodies have produced, maintained, and released land-cover geodatabases over the two last decades.\\
The exploitation of a given geographical extent or level of representation appears also to be correlated to specific processes. "Text" and \textit{local} extent are key elements for "Validation" and "Updating". Conversely, "Observations" and "Vector" mainly relate to "Mapping", and "Observations" basically consider all slave sources as "Validation". "Modelling" is prominent for the \textit{global} extent", \IE the farthest process (spatially speaking) from mapping and validation issues.

%%%%%%%%%%%%%%%%%%%%%%%%%%%%%%%%%%%%%%%%%%%%%%%%%%%%%%%%%%%
%%%%%%%%%%%%%%%%%%%%%%%%%%%%%%%%%%%%%%%%%%%%%%%%%%%%%%%%%%%
\section{Main lines of research}\label{sec:perspectives}

We do not enumerate and discuss on current methodological challenges related to multi-modal geospatial data processing, extensively discussed in the literature \cite{isprs-archives-XLIII-B2-2020-703-2020} \cite{Koldasbayeva} \cite{DevisAI} \cite{Ribana_GRSM25} and on the limited efficiency of machine learning research behind most approaches today \cite{pmlr-v235-karl24a}. We rather focus on key perspectives or missing works for \textit{identically processed sources} from distinct communities and favour retroaction loops between levels of representations and geographical extents.

\subsection{Extending AI models.}
First, among the well-known AI generic model-centric issues, few can be adopted here:
%\begin{itemize}[leftmargin=*]
    %\item
    
\faAngleRight~\textit{Handling unseen sources}: a pivotal challenge is the ability to infer a model with unseen data during the training stage, often coped with fine-tuning solutions \cite{DEBUYSERE202693}. This would ease the adoption of such models to new end-users and their feed with expert/local knowledge. Current effort is made with self-supervised AI collaborative models, mostly with EO images, with co-learning \cite{Colearning}, distillation \cite{8764498}, modality-agnostic or modality-incomplete learning strategies \cite{zhang2023learningunseenmodalityinteraction}. Effort should also be put on Test-Time Training strategies \cite{sun2020testtimetraining} for inherent distribution shift handling.
    %\item
    
    \faAngleRight~\textit{Federated learning}: it targets to train an AI model with data from multiple devices or infrastructures without centralization \cite{DevisAI}. This subject becomes of utmost importance with the multiplication of national or local research infrastructures and challenges in national governance preservation \cite{Research_Infrastructure}. In the same flavour, Mixture-of-Experts strategies are perfectly tailored to handle multiple heterogeneous data sources \cite{MoE}.
    %\item
    
    \faAngleRight~\textit{Knowledge-based learning}: inserting domain- or community-based knowledge is of utmost importance, in particular the formal representations proposed by ontologies \cite{ZHAO20241855}. This bridges the gap between image-based communities with their other computer sciences and GIS counterparts.
    %\item
    
    \faAngleRight~\textit{Handling domain shifts, domain generalization issues} and in particular covariate shifts, \EG with adaptive sensing solutions \cite{baek2025aisensebetterjust} with specific objectives, instead of scaling existing architectures and training data. Mixture-of-Experts and other domain expertise combination strategies are also suitable \cite{CODEX} \cite{lee2026gram}.
    %\item
    
    \faAngleRight~\textit{Mitigating bias transfer} when adopting large geospatial AI models \cite{Ramos_2025_ICCV}: \EG world models inevitably come with pre-training bias that should addressed. Continual pretraining is a solution \cite{mendieta2023towards} along with more real-world benchmarks.
    %\item
    
    \faAngleRight~\textit{Estimating uncertainty and moving to interpretability}. Relevant explainable AI solutions should target to develop interpretable models instead of focusing on deriving ad-hoc mid-architecture outputs from existing AI models \cite{RudinNatureXAI}.

%\end{itemize}

\subsection{Real-world benchmarking and validation.} Heavily related to the previous point, key aspects are:
%\begin{itemize}[leftmargin=*]
%    \item

\faAngleRight~\textit{Consistent and muti-faceted benchmarks}: existing ones rarely capture source and environmental variations, logical dependency between processes \cite{zou2025unimmmu}, as well as stakeholders and other community expectations. Discussions across disciplines should be initiated in this direction \cite{predbench}.

%    \item
\faAngleRight~\textit{Spatially and temporally consistent reference data across data sources and extents for validation}. Such reference is not always reliable, missing in most challenging environments and at very large scales or outdated. Data curation of fine-tuning training data can be also an efficient solution for bias transfer issues \cite{wang2023ICCV}.

%    \item
\faAngleRight~\textit{Handling genuine applications} (economics, history, social sciences) to help understanding complex cases. This would engage discussions across disciplines on how to build data and how to design non brute-force statistical incremental models \cite{doi:10.1086/723723}. This would favour the adoption of barely investigated sources or combinations of sources, beyond the current VLM perspective \cite{GE2025146}.     
%\end{itemize}

\subsection{Adopting a user-centric perspective.} Pure predictive performance on mainstream data sources is over, which calls for:
%\begin{itemize}[leftmargin=*]
    %\item
    
\faAngleRight~\textit{Disantangling data producer, model designer, and end-user perspectives.} Most communities are multi-sources. Yet, researchers remain focused on producing datasets for challenging existing machine learning models or stimulating research on their own communities, driven by maximizing well-known metrics. This does not call data and models into question and does not open the field for retroaction loops. Is my data adapted to my problem? Is my model correctly leveraging my data? Can my model outputs be beneficial for my input data or other data sources? New benchmarks and interdisciplinary initiatives are highly recommended %welcome to bridge the gap between end-user expectations and a given model or data source 
\cite{10.1145/3593013.3593978}.

    %\item
    \faAngleRight~\textit{Adopting a critical analysis of the sources or a hermeneutics perspectives}. Opening data and models is useful. Yet the genuine potential is not correctly assessed: it requires their adoption in other communities to develop an objective assessment of the data and models we produce on a daily basis, and more adapted metrics on what should be maximized \cite{Hermeneutics}.
    
    %\item
    \faAngleRight~\textit{Quo vadis human-centric machine learning?} It has emerged as the research field that investigates the methods of aligning ML models with human "goals", merging the challenges of explainability, interpretability, privacy, etc. \cite{tang2025humancentricfoundationmodelsperception}. The key question is whether such paradigm fits to the geospatial domain which is less restrictive than \EG computer vision \cite{khirodkar2024sapiens}.
    
    %\item
    \faAngleRight~\textit{Enforcing source diversity.} Among the sources listed in Section~\ref{sec:sources}, geospatial numerical traces and text sources remain barely explored, due to their heterogeneous quality and lack of representativeness \cite{ArribasBel_accidental}. However, for most of the challenges mentioned above, they should no longer be overlooked and favored with respect to standard "additional modalities", namely weather data, elevation, or land-cover maps.
    %\end{itemize}

%Je remarque que mon paragraphe est sur les problèmes; tu voulais de perspectives :) EXACTEMENT, DES SOLUTIONS !!
%Lack of spatially and temporally consistent reference data across data sources and extents for validation. Except for a few cases (e.g., global reference data for LC at 10m resolution), reliable reference data are often missing for large areas or are outdated. Indeed, existing reference data, where exist, are rarely updated at the same frequency as data sources (e.g. observation data updated every 5 days), national LULC updated annually, buildings and road networks updated every three months). As a consequence, When multi-sources data are combined significant temporal mismatches arise. These issues become even more significant for change detection processes, no matter the type of data (e.g. building change, LU or LC vector or raster, Lidar HD change), because the references and observations datasets are collected at different times or under evolving specifications. As a result, it becomes difficult to determine and validate real-world changes from changes coming from inconsistencies between the datasets.   

\subsection{Discoverability, and reuse of existing research.} Open models and data does not suffice to comply with FAIR principles \cite{HU2025104477}.

%\begin{itemize}[leftmargin=*]
%    \item
\faAngleRight~\textit{Closing the gap in the discoverability and comparability of available benchmarks, models, and algorithms across research communities}. Much progress is still needed, even in context where FAIR principles are applied, especially by developing ontologies and knowledge graphs to reduce the semantic gap between communities \cite{Arvor18082019}.

%    \item
\faAngleRight~\textit{Improving the tools for providing curating and comparable research resources, and user feedbacks on such resources}. This includes papers, data, algorithms with descriptive and a critical analysis \cite{bucher2025}. Agentic AI can be adopted in order to organize, annotate, and suggest relevant resources \cite{siebenmann2026ogd4all}. 
    
%\end{itemize}

%%%%%%%%%%%%%%%%%%%%%%%%%%%%%%%%%%%%%%%%%%%%%%%%%%%%%%%%%%%
%%%%%%%%%%%%%%%%%%%%%%%%%%%%%%%%%%%%%%%%%%%%%%%%%%%%%%%%%%%
\section{Conclusion}\label{sec:conclusion}
In a native multi-source world, a key issue is to adequately and optimally exploit all available geospatial, even rare, sources. Community and discipline bias exist and siloed reasoning has dominated for numerous years. It results in a privileged adoption of specific sources for specific processes, without retroaction loops. At the same time, the current predominance of AI models lead to a more and more agnostic approach of data fusion problems, shading the particularity of each source. We advocate that an in-between strategy, considering all data sources at the same level, would be beneficial to leverage all sources. The geospatial-related communities should therefore devote more research efforts in barely documented fusion configurations and how to take advantage of main outputs of other communities or disciplines for our own key applications.

%%%%%%%%%%%%%%%%%%%%%%%%%%%%%%%%%%%%%%%%%%%%%ùù
\section*{Acknowledgments}
The authors would like to thank Julien Perret, Nicolas Gonthier, and Cédric Véga for fruitful discussions on the subject.

%%%%%%%%%%%%%%%%%%%%%%%%%%%%%%%%%%%%%%%%%%%%%%%%%%%%%%%%%%%
{
	\begin{spacing}{0.99}
		\normalsize
		\bibliography{biblio_global-local} % Include your own bibliography (*.bib), style is given in isprs.cls
	\end{spacing}
}

\end{document}